\documentclass[10pt,twocolumn,letterpaper]{article}

\usepackage{cvpr}              %

\usepackage{graphicx}
\usepackage{amsmath}
\usepackage{amssymb}
\usepackage{booktabs}
\usepackage{svg}

\usepackage{comment}
\usepackage{bbding}
\usepackage{makecell}
\usepackage{multirow}
\usepackage{tabularx}
\usepackage[page]{appendix} %

\usepackage{titletoc}
\usepackage[accsupp]{axessibility}

\usepackage{algorithm}
\usepackage{algorithmic}

\usepackage[pagebackref,breaklinks,colorlinks]{hyperref}

\usepackage[capitalize]{cleveref}
\crefname{section}{Sec.}{Secs.}
\Crefname{section}{Section}{Sections}
\Crefname{table}{Table}{Tables}
\crefname{table}{Tab.}{Tabs.}

\usepackage{xcolor}
\usepackage{enumitem}
\usepackage{xspace}
\usepackage{booktabs}
\usepackage{colortbl}
\usepackage{multirow}
\usepackage{makecell}
\usepackage{lipsum} %
\usepackage{gensymb} %

\newcommand{\Tref}[1]{Table~\ref{#1}}
\newcommand{\eref}[1]{Eq.~\eqref{#1}}

\newcommand{\fref}[1]{Fig.~\ref{#1}}
\newcommand{\Fref}[1]{Figure~\ref{#1}}

\usepackage[labelsep=period]{caption}
\renewcommand{\paragraph}[1]{\vspace{0.2em}\noindent \textbf{#1 \hspace{0.2em}}}

\newcommand{\hashgrid}{hash-grid}

\newcommand{\zyqm}[1]{\textcolor{black}{{#1}}}

\definecolor{MyDarkRed}{rgb}{0.66, 0.16, 0.16}
\definecolor{MyDarkBlue}{rgb}{0.16, 0.16, 0.66}

\newcommand{\semanticfusion}{scale-adaptive semantic label fusion\xspace}
\newcommand{\semanticfusionshort}{scale-adaptive fusion\xspace}
\newcommand{\SemanticFusion}{Scale-adaptive Semantic Label Fusion\xspace}
\newcommand{\instancefusion}{cross-view instance label grouping\xspace}
\newcommand{\instancefusionshort}{cross-view grouping\xspace}
\newcommand{\InstanceFusion}{Cross-view Instance Label Grouping\xspace}
\newcommand{\detectronlabel}{Detectron-Label\xspace}
\newcommand{\SAMlabel}{SAM-Label\xspace}

\def\confName{CVPR}
\def\confYear{2024}

\begin{document}

\title{Aerial Lifting: Neural Urban Semantic and Building Instance Lifting \\ from Aerial Imagery}

\author{Yuqi Zhang$^{1,2}$ \quad Guanying Chen$^{1,3}$\thanks{Corresponding author} \quad Jiaxing Chen$^{1,3}$ \quad Shuguang Cui$^{2,1}$ \vspace{0.3em} \\
{\normalsize $^1$FNii, CUHKSZ}
\quad{\normalsize $^2$SSE, CUHKSZ} \quad {\normalsize $^3$Sun Yat-sen University}
}

\maketitle

\begin{abstract}
    We present a neural radiance field method for urban-scale semantic and building-level instance segmentation from aerial images by lifting noisy 2D labels to 3D. This is a challenging problem due to two primary reasons. Firstly, objects in urban aerial images exhibit substantial variations in size, including buildings, cars, and roads, which pose a significant challenge for accurate 2D segmentation. Secondly, the 2D labels generated by existing segmentation methods suffer from the multi-view inconsistency problem, especially in the case of aerial images, where each image captures only a small portion of the entire scene. To overcome these limitations, we first introduce a \semanticfusion strategy that enhances the segmentation of objects of varying sizes by combining labels predicted from different altitudes, harnessing the novel-view synthesis capabilities of NeRF. We then introduce a novel \instancefusion based on the 3D scene representation to mitigate the multi-view inconsistency problem in the 2D instance labels. Furthermore, we exploit multi-view reconstructed depth priors to improve the geometric quality of the reconstructed radiance field, resulting in enhanced segmentation results. Experiments on multiple real-world urban-scale datasets demonstrate that our approach outperforms existing methods, highlighting its effectiveness. The source code is available at \href{https://github.com/zyqz97/Aerial\_lifting}{https://github.com/zyqz97/Aerial\_lifting}.
\end{abstract}

\section{Introduction}
\label{sec:intro}

3D urban-scale semantic understanding plays a crucial role in various applications, from urban planning to autonomous driving systems. 
Accurate semantic and instance-level segmentation of objects in 3D scenes is essential for a wide range of tasks. 

Existing 3D urban semantic understanding methods primarily rely on point cloud representation \cite{qi16pointnet,hu2020randla}. They typically train a point cloud segmentation method on labeled 3D datasets \cite{hu2021towards}. However, annotating 3D data is labor-intensive, posing challenges in creating a comprehensive training dataset with diverse scenes. 

\begin{figure}[tb] \centering
    \includegraphics[width=0.48\textwidth]{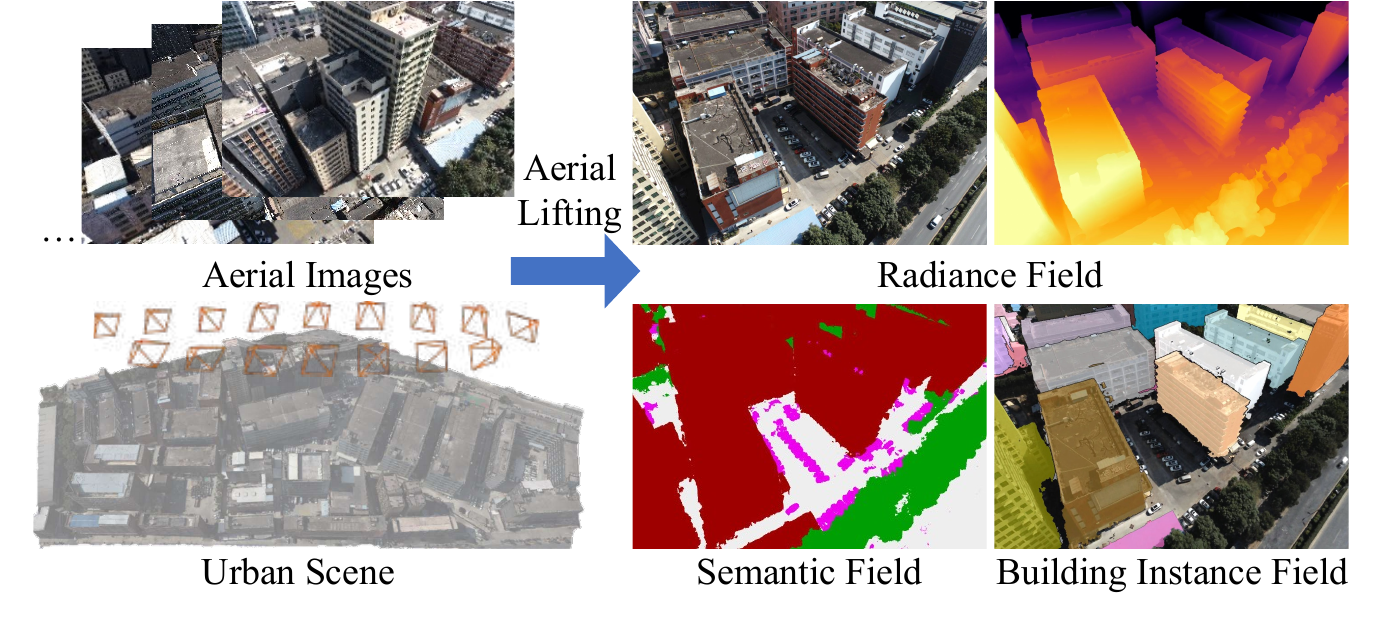}
    \caption{Given multi-view aerial images, our method lifts 2D labels to optimize the radiance, semantic, and instance fields for urban-scale semantic and building-level instance understanding.} \label{fig:teaser}
\end{figure}

Recently, neural radiance fields (NeRF) \cite{mildenhall2020nerf} have emerged as an effective 3D scene representation, enabling photorealistic rendering of fine details.
Several methods are proposed to perform semantic segmentation or panoptic segmentation on NeRF by lifting 2D estimation to 3D \cite{zhi2021place,siddiqui2023panoptic}. However, these methods mainly validate on the room-scale indoor scenes or street-view outdoor scenes.
In this work, we aim to perform urban-scale semantic and building-level instance segmentation from multi-view aerial images. 
Our method leverages neural radiance fields to lift noisy 2D labels to a 3D representation without manual 3D annotations, effectively bridging the gap between 2D imagery and the complex 3D urban environment (see~\fref{fig:teaser}). 

This is inherently challenging due to several factors.
On one hand, urban aerial images capture scenes that encompass a wide range of object sizes, including buildings, vehicles, and roads \cite{lyu2020uavid}. 
Existing segmentation methods often struggle to handle these variations effectively as their training data distribution is different from that of aerial images \cite{cheng2022masked}, or lack large-scale labeled aerial images for fine-tuning \cite{wang2022unetformer}.
On the other hand, 2D instance labels generated by existing segmentation methods often suffer from the multi-view inconsistency problem (\eg, an object is segmented as one instance in a view might be segmented into multiple independent instances in another view). This problem becomes particularly pronounced in the context of aerial images, where each image captures only a small portion of the entire scene. 
Furthermore, the geometry reconstruction quality of the large-scale scene will largely affect the semantic segmentation.

To address these problems, we introduce three key strategies to enhance the accuracy and robustness of our segmentation approach.
First, we propose a \emph{\semanticfusion} strategy, enabling the segmentation of objects of varying sizes by fusing labels predicted from different altitudes. This leverages the novel-view synthesis capabilities of NeRF \cite{mildenhall2020nerf} to render photorealistic images at different altitudes. 
Second, we introduce a \emph{\instancefusion} strategy to group instance labels in a view utilizing information from other views. It is achieved by performing cross-view label projection based on the relative camera poses and geometry of the 3D scene representation. 
This strategy effectively mitigates the multi-view inconsistency problem in the 2D instance labels, providing a more coherent and accurate segmentation of urban objects. 
Furthermore, we exploit \emph{depth priors obtained from multi-view stereo} to improve the geometric quality of the reconstructed radiance field, ultimately leading to enhanced segmentation results. Our approach has been extensively evaluated on multiple real-world urban-scale \zyqm{scenes}, demonstrating its superior performance compared to existing methods.

In summary, the key contributions are as follows:
\begin{itemize}[itemsep=0pt,parsep=0pt,topsep=2bp]
    \item We present a novel radiance field approach for urban-scale semantic and building-level instance segmentation from aerial images by lifting noisy 2D labels to 3D, achieving state-of-the-art results.
    \item We introduce a \semanticfusion strategy that combines 2D labels predicted from different altitudes to enhance the segmentation of objects of varying sizes, leveraging NeRF's novel-view synthesis capabilities.
    \item We present a \instancefusion approach based on the 3D scene representation to mitigate the multi-view inconsistency problem in 2D instance labels, resulting in more reliable instance segmentation results.
\end{itemize}

\section{Related Work}
\label{sec:related_works}

\paragraph{3D Urban Semantic Learning}
Traditional 3D semantic learning methods involve training models on 3D datasets with ground-truth annotations \cite{qi2017pointnet++,lai2022stratified,graham20183d,choy20194d,yi2019gspn,yang2019learning,vu2022softgroup,hurtado2020mopt,hong2021lidar,mao2022beyond,hu2022sensaturban}. These methods often operate on explicit representations such as point clouds \cite{qi16pointnet,hu2020randla}. 
For 3D urban scenes, some methods perform 3D building instance segmentation from meshes \cite{chen20223,adam2023deep,blaha2016large} or point clouds \cite{yang2023urbanbis,nguatem2017modeling,chen2022stpls3d}.
Recent research shows that implicit representations \cite{park2019deepsdf,yariv2020multiview,niemeyer2020differentiable} can effectively represent continuous and detailed surfaces and enable differentiable rendering, making them a promising choice for semantic understanding. In this work, we leverage the radiance field representation \cite{mildenhall2020nerf} and lift the estimated 2D labels to 3D through per-scene optimization.

\paragraph{Neural Scene Representations}
Traditional multi-view 3D reconstruction methods \cite{agarwal2011building,fruh2004automated,li2008modeling,zhu2018very,snavely2006phototourism,UrbanScene3D} often apply structure-from-motion (SFM) techniques to estimate camera poses \cite{schonberger2016_sfm_cvpr16}, followed by dense multi-view stereo \cite{furukawa2010towards,furukawa2010pmvs} to generate 3D models. 

Recently, neural scene representations have achieved significant success in 3D scene modeling~\cite{shum2000review,tewari2020state,tewari2021advances,xie2022neural, torchngp, tang2022compressible}.
Specifically, neural radiance fields (NeRF)~\cite{mildenhall2020nerf} achieve photorealistic rendering for diverse scenes. 
Subsequently, many methods have been proposed to enhance NeRF in various aspects, including surface geometry \cite{oechsle2021unisurf,wang2021neus,yariv2021volume,yu2022monosdf} and optimization speed \cite{sun2022direct,fridovich2022plenoxels,muller2022instant,chen2022tensorf,kerbl20233d}.
To handle large-scale scenes, several methods introduce sophisticated designs to improve rendering quality and reconstruction geometry \cite{guo2023streetsurf,xiangli2022bungeenerf,zhang2022nerfusion,wu2022scalable,zhang2023efficient,xu2023grid,zhenxing2022switch,li2023matrixcity,lu2023large,mari2022sat}. For example, Block-NeRF \cite{tancik2022block} and Mega-NeRF \cite{turki2022mega} decompose the scene into several partitions, with each partition represented by a different local NeRF. 
\zyqm{StreetSurf~\cite{guo2023streetsurf} proposed a variant of the \hashgrid~\cite{muller2022instant}, which allocates the grid space according to the ratios of the three axes, making full use of the grid space.}
In this work, our goal is to extend NeRF to achieve urban-scale semantic understanding.

\begin{figure*}[tb] \centering
    \includegraphics[width=\textwidth]{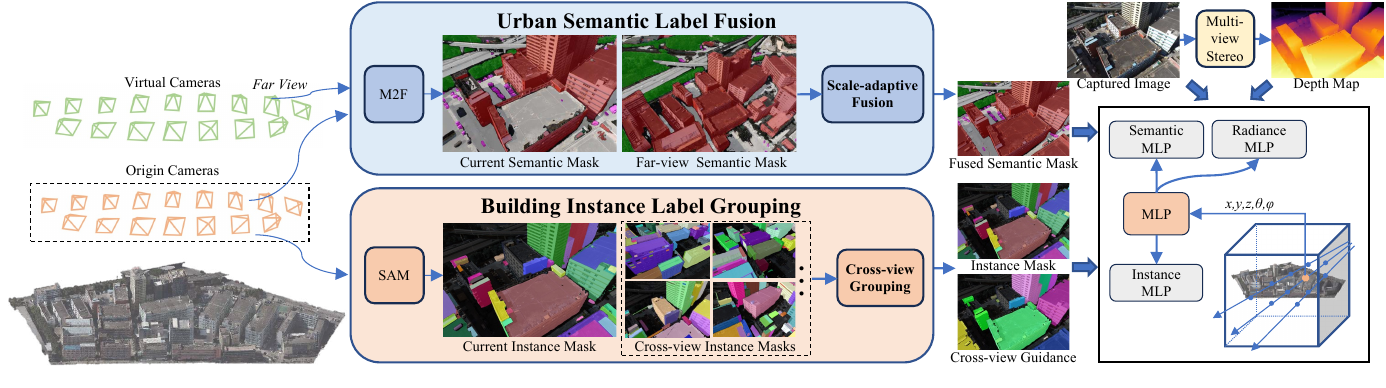}
    \caption{Overview. We present a neural radiance field (NeRF) method for urban-scale semantic and building-level instance segmentation from aerial images by lifting noisy 2D labels to 3D.
    For semantic segmentation, we adopt a \emph{\semanticfusion} strategy to fuse the semantic labels from different altitudes using images rendered by NeRF, mitigating the ambiguities of the 2D semantic labels.
    For instance segmentation, we propose a \emph{\instancefusion} strategy to guide the training of instance field. 
    In addition, \zyqm{a depth prior} from Multi-view Stereo (MVS) is introduced to enhance the geometry reconstruction, leading to more accurate semantic learning.
    } \label{fig:overview}
\end{figure*}

\paragraph{Semantic Understanding with Neural Fields}
Recent research has explored the use of NeRF for semantic understanding of 3D scenes. Semantic-NeRF \cite{zhi2021place} fuses 2D semantic labels into 3D using an additional MLP branch to predict semantic logits \cite{vora2021nesf,mascaro2021diffuser}.
A similar idea of fusing 2D to 3D with NeRF has also been applied to fuse multi-view features \cite{kerr2023lerf,kobayashi2022decomposing,hong20233d,takmaz2023openmask3d,blomqvist2023neural} to enable open-vocabulary understanding.
Moreover, leveraging the powerful segment anything (SAM) model \cite{kirillov2023segany}, SA3D \cite{cen2023segment} proposes to segment a single object in NeRF with a user click \cite{chen2023interactive}. 

For 3D panoptic segmentation, one of the main challenges is obtaining appropriate instance supervision across multiple views \cite{hu2023instance,wang2022dm}.
Instance-NeRF \cite{hu2023instance} and Panoptic NeRF \cite{fu2022panoptic} use 3D instance information for training.
PNF \cite{kundu2022panoptic} relies on object tracking to provide instance supervision.
Panoptic Lifting \cite{siddiqui2023panoptic} adopts linear assignment to match the current predicted 3D instance with the provided 2D labels. 
Contrastive Lifting \cite{bhalgat2023contrastive} utilizes feature contrastive learning, followed by clustering to obtain instance information \cite{cheng2023panoptic}.
However, existing methods are mainly designed for indoor \cite{roberts2021hypersim,dai2017scannet,straub2019replica} or outdoor street-view \cite{liao2022kitti,geiger2012we} scenes. 
In contrast, our focus is on urban scene understanding from aerial images, which is particularly challenging as aerial images encompass a wide range of object sizes and each image captures only a small portion of the entire scene.

\newcommand{\viewdir}{\boldsymbol{d}}
\newcommand{\pointcolor}{\boldsymbol{c}}
\newcommand{\pointdepth}{\boldsymbol{t}}

\newcommand{\density}{\sigma}
\newcommand{\ray}{\boldsymbol{r}}
\newcommand{\rayset}{\boldsymbol{R}}
\newcommand{\pixelcolor}{C}
\newcommand{\pixeldepth}{D}
\newcommand{\coarse}{{(c)}}
\newcommand{\refine}{{(r)}}
\newcommand{\loss}{\mathcal{L}}
\newcommand{\gt}{\mathrm{(gt)}}
\newcommand{\uncertain}{{(\tau)}}
\newcommand{\point}{\boldsymbol{x}}
\newcommand{\imgset}{\mathbf{I}}
\newcommand{\msemanticset}{\mathbf{M}}
\newcommand{\minstanceset}{\mathbf{H}}
\newcommand{\msam}{\mathbf{H}}
\newcommand{\nview}{M}
\newcommand{\viewindx}{N_p}
\newcommand{\nlight}{L_m}
\newcommand{\camloc}{\boldsymbol{o}}
\newcommand{\semantic}{\boldsymbol{s}}
\newcommand{\transform}{\mathbf{T}}
\newcommand{\depthmap}{\tilde{\pixeldepth}}

\section{Method}%

\subsection{Overview}
Given multi-view posed aerial images $\{\imgset \}$ of an urban scene, we perform 3D semantic and building-level instance understanding of the scene based on the neural radiance field (NeRF) \cite{mildenhall2020nerf}.
Our method applies off-the-shelf methods \cite{cheng2022masked,kirillov2023segany} to obtain the semantic labels $\{\msemanticset \}$ and instance labels $\{\minstanceset \}$ for input images, and then lifts the noisy 2D labels to 3D via per-scene optimization (see \fref{fig:overview}). 

\begin{figure}[t] \begin{center}
    \makebox[0.155\textwidth]{\footnotesize Input}
    \makebox[0.155\textwidth]{\footnotesize UNetFormer}
    \makebox[0.155\textwidth]{\footnotesize Mask2Former}
    \\
    \includegraphics[width=0.155\textwidth]{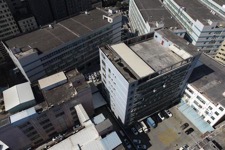}
    \includegraphics[width=0.155\textwidth]{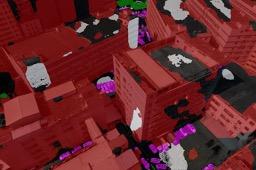}
    \includegraphics[width=0.155\textwidth]{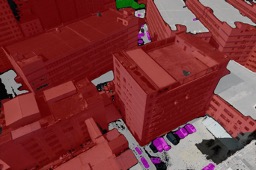}
    \\
    \vspace{-2pt} 
    \makebox[0.48\textwidth]{\small (a) Semantic segmentation from 2D methods}
    \\
    \vspace{1pt} 
    \makebox[0.155\textwidth]{\footnotesize Input}
    \makebox[0.155\textwidth]{\footnotesize Detectron}
    \makebox[0.155\textwidth]{\footnotesize SAM}
    \includegraphics[width=0.155\textwidth]{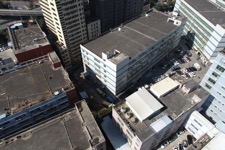}
    \includegraphics[width=0.155\textwidth]{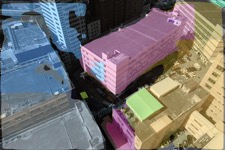}
    \includegraphics[width=0.155\textwidth]{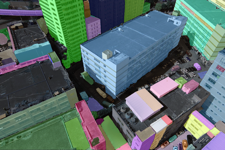}
    \\
    \vspace{-2pt} 
    \makebox[0.48\textwidth]{\small (b) Instance segmentation from 2D methods}
    
    \caption{Problem of existing 2D semantic and instance \zyqm{segmentation} methods. (a) We use the red color to highlight the buildings and white for roads. 
    UNetFormer suffers from recognizing road and Mask2Former suffers from misclassification between rooftops and roads. (b) Distinctive colors are assigned to different instances. The instance labels obtained from Detectron appear overly large, while those from SAM seem excessively small.} \label{fig:semantic_seg}
    \vspace{-1.05em}
\end{center} \end{figure}

\paragraph{Challenges}
There are two critical challenges that need to be addressed.
First, due to significant variations in object size, state-of-the-art semantic segmentation methods, such as Mask2Former \cite{cheng2022masked, cheng2021maskformer} trained on daily images and UNetFormer \cite{wang2022unetformer} trained on a small scale of aerial images, struggle to generate reliable semantic labels for aerial images (see \fref{fig:semantic_seg}~(a)).
Secondly, obtaining accurate building instance segmentation is challenging due to the diverse shapes and substantial size of buildings.
\Fref{fig:semantic_seg}~(b) shows that the leading method \cite{he2017mask,spacenet2023} for building instance segmentation in aerial images \zyqm{fails} to generate robust instance labels, especially for dense cluster of buildings. 
Recently, SAM~\cite{kirillov2023segany} demonstrates superior generalization ability in semantic-agnostic instance segmentation with precise mask boundaries.
However, SAM produces over-segmented masks and multi-view inconsistent instance segmentation (\eg, a building segmented as one instance in one view might become multiple different instances in other views).

Lifting such inaccurate 2D labels to 3D with NeRF results in inaccurate 3D semantic and instance segmentation.
To address these issues, we introduce a \emph{\semanticfusion} strategy for semantic segmentation and a \emph{\instancefusion} strategy for instance segmentation to provide more accurate and consistent supervision for the 2D-to-3D lifting process.

\subsection{3D Scene Representation}

\paragraph{Neural Radiance Field}
We represent the geometry and appearance of a 3D scene with NeRF \cite{mildenhall2020nerf}, which employs a continuous function to map a 3D point $\point_k$ in space and view direction $\viewdir$ to density $\density_k$ and color $\pointcolor_k$.
The pixel color can be computed by integrating the color of the points sampled along its visual ray $\ray$  through volume rendering:
\begin{align}
    \label{eq:volumerender}
    \tilde{\pixelcolor}(\ray) = \sum_{k=1}^K T_k(1 - \exp(-\density_k\delta_k))\pointcolor_k, 
\end{align}
where $T_k  = \exp\left(-\sum_{j=1}^{k-1} \density_j\delta_j\right)$, and $\delta_{k}=t_{k+1}-t_{k}$ is the distance between adjacent sampled points. %
During optimization, a NeRF is fitted to a scene by minimizing the reconstruction error between the rendered color $\tilde{\pixelcolor}$ and the captured color $\pixelcolor$ in the sampled ray set $\rayset$:
\begin{equation}
    \label{eq:color_loss}
    \mathcal{L}_\text{color} = \sum_{\ray \in \rayset}\| \tilde{\pixelcolor} (\ray) - \pixelcolor (\ray) \|_2^2.
\end{equation}

\paragraph{Semantic and Instance Fields}%
We follow semantic-NeRF~\cite{zhi2021place} and panoptic lifting \cite{siddiqui2023panoptic} to add a semantic branch and an instance branch to represent the 3D semantic and instance fields.
The semantic and instance labels $\tilde{S}(\ray)$ of a ray can be rendered by volume rendering as \eref{eq:volumerender}:
\begin{equation}
    \tilde{S}(\ray) = \sum_{k=1}^K T_k(1 - \exp(-\density_k\delta_k))\semantic_k
\end{equation}
where $\semantic_k$ is the semantic or instance output of a 3D point. 

Given the 2D semantic labels for each image, the semantic field can be optimized by minimizing the multi-class cross-entropy loss $\mathcal{L}_\text{semantic}$ between the rendered semantic labels and the 2D labels \cite{zhi2021place}.
The loss function for instance field $\mathcal{L}_\text{instance}$ needs special design, as the instance IDs of the same 3D instance predicted from different images are not consistent (\eg, a building instance might have an ID of $1$ in one view and an ID of $2$ in another view). Existing methods propose to solve a linear assignment problem to match the best 3D and 2D instance pairs \cite{siddiqui2023panoptic} or utilize contrastive feature learning to cluster 3D instances \cite{bhalgat2023contrastive}. 
\emph{Note that these methods do not consider the problem of multi-view instance label inconsistency, where an instance might be segmented into multiple different instances in different views, which is common in urban aerial images.}

\subsection{Urban Semantic Label Fusion}

We employ the state-of-the-art Mask2Former \cite{cheng2022masked} to estimate 2D segmentation masks $\{\msemanticset\}$ for input views, focusing on the four primary categories in the urban landscape: \emph{Buildings}, \emph{Trees}, \emph{Cars}, and \emph{Roads}. 
Other methods can also be used, but we found Mask2Former is more robust and accurate \cite{siddiqui2023panoptic,bhalgat2023contrastive}.
However, segmentation labels generated through 2D methods suffer from ambiguities, \eg building rooftops may erroneously be labeled as road surfaces (see \fref{fig:semantic_seg}~(a)). This misclassification stems from the scale variability inherent in aerial imagery, where each image captures only a limited portion of the large building.

\begin{figure}[t] \begin{center}
    \includegraphics[width=0.48\textwidth]{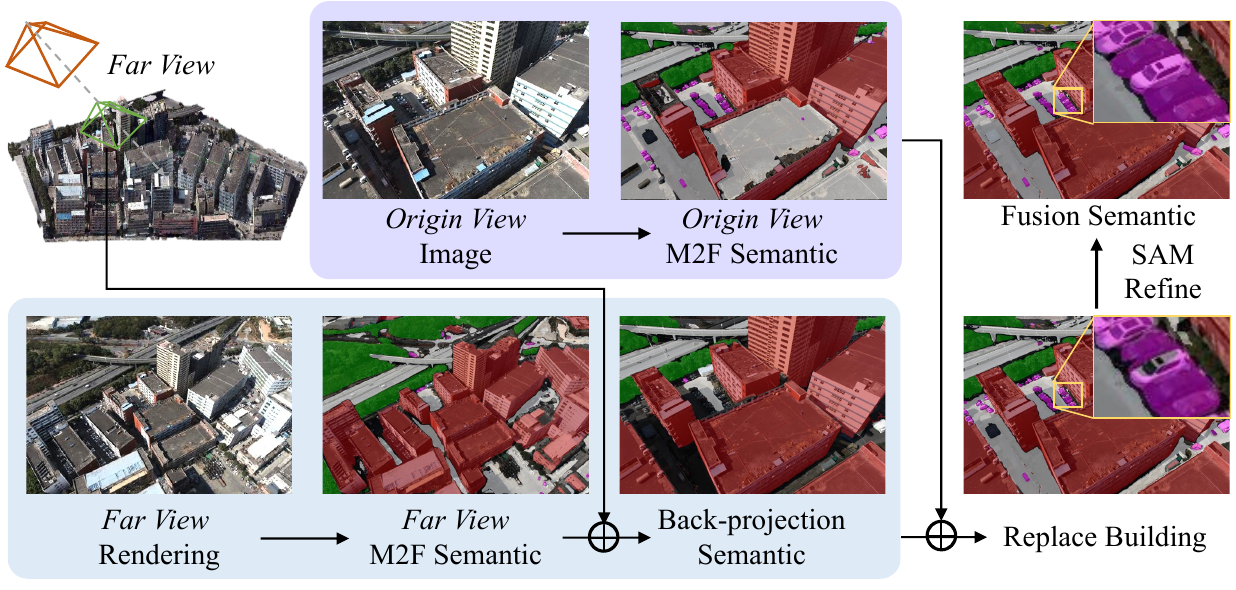}
    \\
    \vspace{-5pt}
    \makebox[0.48\textwidth]{\footnotesize (a) Illustration of \SemanticFusion}
    \\
    \vspace{6pt}
    \includegraphics[height=0.14\textwidth,width=0.21\textwidth]{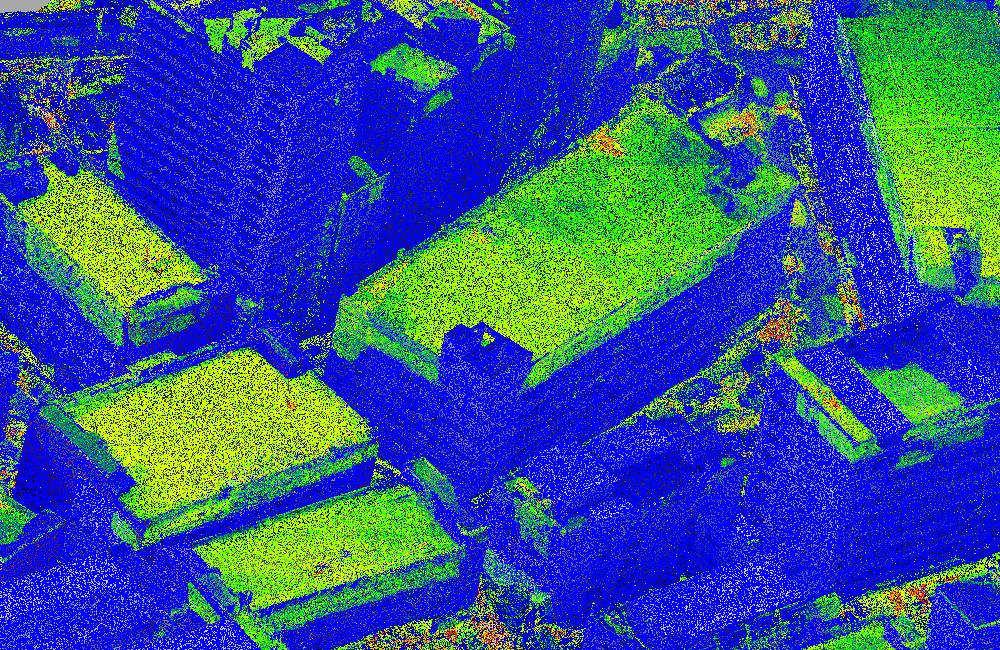}
    \includegraphics[height=0.14\textwidth,width=0.21\textwidth]{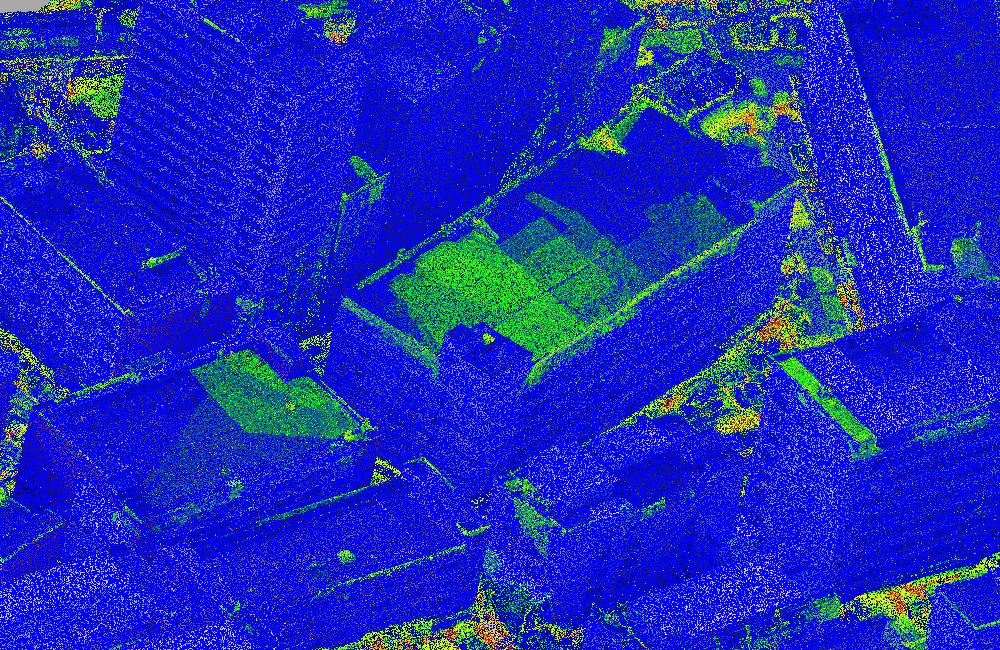}
    \includegraphics[height=0.14\textwidth,width=0.03\textwidth]{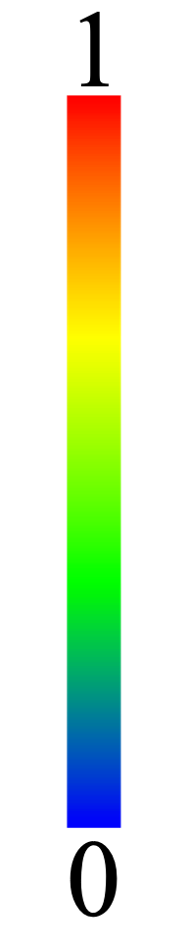}
    \\
    \vspace{-4pt}
    \makebox[0.2\textwidth]{\footnotesize (b) Label Conflict in M2F}
    \makebox[0.2\textwidth]{\footnotesize (c) Label Conflict in Ours}
    \makebox[0.03\textwidth]{}
    \\
    \caption{(a) Illustration of the \semanticfusion process. (b)-(c) Visualization of the conflict level in 3D points using entropy, where higher entropy indicates higher conflict level.
    } 
    \label{fig:semantic_fusion}
    \end{center} 
\end{figure}

To circumvent this issue, we propose a \semanticfusion strategy to improve the semantic label. 
This idea stems from the observation that the semantic labeling of large object categories (\eg, \emph{building}) is more reliable when viewed from a distance view point, as the object will become smaller in the context (see \fref{fig:semantic_fusion}). 

\paragraph{\SemanticFusion}
NeRF has a great ability for photorealistic novel-view synthesis compared to the explicit representations, \eg, point clouds. 
We perform novel-view synthesis based on the radiance field to simulate images captured from elevated altitudes. For each \zyqm{original} image, we increase its camera altitudes for novel-view rendering, rendering a set of far view images $\{\imgset^\text{f}\}$. 
We then compute the segmentation mask $\{\msemanticset^\text{f}\}$ for the far view images.
By leveraging the depth information derived from the neural radiance field, the segmentation obtained from the \emph{far view} is then back-projected to the original view for refining the mask of the building category. 
Specifically, considering a pixel with the coordinate of $p^f$ in a far view image, the projected pixel coordinate $p^o$ in the original image is defined as:
\begin{align}
    \label{eq:project}
    p^o \sim K \transform_{f\to o} \depthmap^f(p^f) K^{-1} p^f,
\end{align}
where $K$ is the camera intrinsic, $\transform_{f\to o}$ is the relative transformation from the far to the original camera, and $\depthmap^f$ represents the rendered depth map of the far view image. 

Furthermore, we apply SAM on the original captured images to predict semantic agnostic masks, which will be utilized to refine the masks of small-scale categories, \eg \emph{trees} and \emph{cars}. 
Specifically, for each semantic mask of the small-scale categories, we match it with the SAM mask that has an intersection of union (IoU) larger than $0.5$. The matched SAM masks will be the refined semantic mask.

To verify the effectiveness of our method, we measure the label consistency in 3D based on the UrbanBIS dataset~\cite{yang2023urbanbis}. 
Given the 3D point cloud, we back-project the predicted 2D semantic label for each view to the 3D space leveraging the camera poses.
Each 3D point will receive multiple 2D semantic labels from different views, and we compute the entropy to measure the inconsistency across views, reflecting the accuracy of labels.  
\Fref{fig:semantic_fusion} presents the conflict with entropy and shows the visualization results. Compared to Mask2Former, our scale-adaptive integration for semantic labels significantly reduces ambiguity between building rooftops and roadways at the original scale, thereby improving per-view segmentation accuracy and essentially reducing the difficulty of 2D-to-3D lifting.

\subsection{Building Instance Label Grouping}

\paragraph{Semantic-agnostic Instance Generation}
Existing instance segmentation methods \cite{he2017mask,spacenet2023} struggle with robust instance segmentation for aerial images of diverse urban scenes.
Impressed by the superior generalization ability of SAM \cite{kirillov2023segany}, we utilize SAM to generate semantic-agnostic masks for building instance segmentation. 
For each image, a grid of \(32 \times 32\) points will be utilized as the input prompt for SAM to predict a set of possible instances.

However, despite its generality, the mask generated by SAM has two characteristics that harm the building instance segmentation: 
1) The SAM model generates masks at different levels of granularity, which might lead to small masks nested inside larger ones, resulting in redundant masks that belong to the same instance  (\eg, window mask on top of the building mask).
2) The generated 2D masks for the same 3D instance are not consistent across multi-view, \eg, a building instance \zyqm{which} is accurately segmented in one view might be segmented into multiple different instances.

\paragraph{Geometry-guided Instance Filtering}
The geometry-guided instance filtering is designed to identify and remove smaller masks nested inside larger masks and exhibit limited height variation.  
Specifically, leveraging the camera parameters and the depth map $\tilde{\pixeldepth}$ of each image computed from the radiance field, we map pixels of each mask to 3D space to determine their physical heights as the difference of the highest and the lowest altitudes. Subsequently, we filter the nested masks with heights smaller than a threshold.

\begin{figure}[t] \begin{center}
    \includegraphics[width=0.48\textwidth]{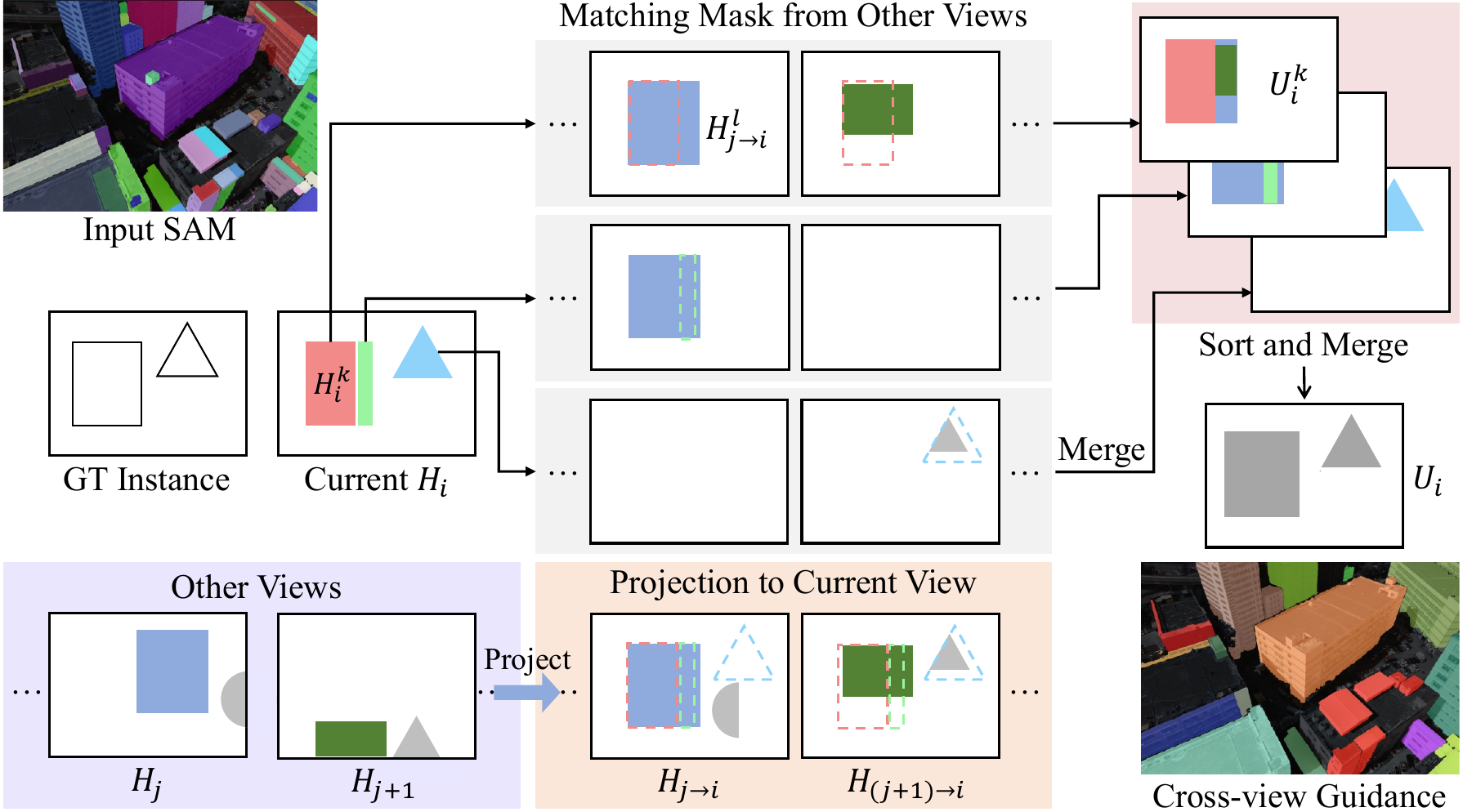}
    \caption{Illustration of \instancefusion. Given the SAM mask of one view, we can get the cross-view guidance map for the instance field training.} \label{fig:crossview_fusion}
    \vspace{-1em}
\end{center} \end{figure}

\paragraph{\InstanceFusion}
As an instance might be segmented into different blocks in different views by SAM, directly lifting SAM masks to the 3D instance field is suboptimal as 3D points will receive conflict supervision in different views. 
To resolve this problem, we introduce a \instancefusion strategy.
The key idea is to synchronize the instance segmentation across different views, thereby consolidating smaller segmented instances into a singular, coherent instance (see \fref{fig:crossview_fusion}).

Consider a scenario with $N$ images. For each image, denoted as the $i$-th view, we have a set of predicted SAM masks, represented as $\msam_i$. When examining the instance segmentation from the perspective of the $i$-th view, it is essential to incorporate the segmentation information from other views. To achieve this, we project the SAM masks from all other views ($j$) onto the $i$-th view. This set of projected masks is denoted by $\{\msam_{j\to i} | j = 1, \dots, N, j \neq i\}$, using camera parameters and depth as specified in \eref{eq:project}.

For each instance mask $\msam_i^k$ in the $i$-th view, we seek to identify corresponding masks in $\msam_{j\to i}$. A match between a pair of masks, $\msam_i^k$ and $\msam_{j\to i}^l$, is established if the intersection-over-minimum-area ratio exceeds a predefined threshold $\tau$ as 
$\frac{|\msam_i^k \cap \msam_{j\to i}^l|}{\min(|\msam_i^k|, |\msam_{j\to i}^l|)} > \tau$,
where $|\cdot|$ represents the area of a mask, \zyqm{and $\tau$ is set to 0.5}. Upon identifying a match, the corresponding masks are merged by uniting their areas, resulting in an expanded mask $\msam_{i\cup j}^k$. This process is repeated for all matches, leading to a collection of expanded masks. These expanded masks are then combined to form a comprehensive \emph{cross-view mask} for each instance as \hbox{$U_i^k = \bigcup_{j \neq i} \msam_{i\cup j}^k$}. 

This procedure is executed for every instance mask in the $i$-th view, resulting into a set of cross-view masks. These masks are then organized in ascending order based on their areas, and the mask value is set to the ID of the instance mask. Subsequently, they are sequentially layered onto a map of dimensions $H \times W$, creating the \emph{cross-view guidance map}, $U_i$. In this map, smaller masks are progressively overwritten by larger ones, which effectively groups the instances more accurately.

With the help of the cross-view guidance map, different instances in the current view are considered the same group if more than $50$\% of their pixels in the cross-view guidance map have the same value. 
During training, we randomly select a single instance from each group.
This approach substantially reduces the occurrence of conflicts in the dense SAM mask annotations, such as when two pixels from the same building instance might be incorrectly classified as belonging to separate instances.

\subsection{Depth Priors from Multi-view Stereo}
In the case of expansive urban environments and sparse observations, optimizing the radiance field solely with the photometric loss can result in imprecise geometry and floating artifacts. 
To mitigate this, our method integrates depth cues derived from multi-view stereo techniques to enforce geometric consistency \cite{schonberger2016_sfm_cvpr16,schonberger2016pixelwise}. 
We reconstruct depth map $\pixeldepth$ for each view and incorporate a depth regularization term in our loss function to refine the NeRF's geometry:
\begin{equation}
    \label{eq:depth_loss}
    \mathcal{L}_\text{depth} = \sum_{\ray \in \rayset} \| \depthmap(\ray) - \pixeldepth(\ray) \| _2^2,
\end{equation}
where $\depthmap(\ray)$ denotes the rendered depth obtained by volume rendering the ray distance in a similar way as in \eref{eq:volumerender}.

Previous studies have leveraged monocular depth \cite{yu2022monosdf} or sparse point information \cite{deng2022depth,fu2022geo} for similar purposes.
 However, we found that the monocular depth estimation methods \cite{ranftl2020towards} are not robust to aerial images, especially for views orthogonal to the ground, while the supervision from the sparse depth information is not sufficient for urban scenes.

\subsection{Optimization}
The overall loss function can be written as:
\begin{equation}
    \label{eq:depth_loss}
    \mathcal{L} = \loss_\text{color} + \lambda_d \mathcal{L}_\text{depth} + \lambda_s \mathcal{L}_\text{semantic} + \lambda_i \mathcal{L}_\text{instance},
\end{equation}
where $\lambda_d$, $\lambda_s$, and $\lambda_i$ are the loss weights and set to 1 in the experiments.

During optimization, we first optimize the radiance field to recover the scene geometry and appearance, and then optimize the semantic and instance fields.
The loss function for semantic is the multi-class cross-entropy loss \cite{zhi2021place}. 
For instance field optimization, we integrate our \instancefusionshort strategy with loss functions introduced by contrastive lifting \cite{bhalgat2023contrastive} and panoptic lifting \cite{siddiqui2023panoptic}. 
During optimization, we filter image rays that do not belong to the building category.
Experiments show that our method effectively improves contrastive lifting and panoptic lifting for building instance segmentation in urban scenes.

\section{Experiments}%
\label{sec:Experiments}
\paragraph{Datasets}
We evaluate our method on the real-world urban scene dataset, named UrbanBIS~\cite{yang2023urbanbis} dataset.
UrbanBIS dataset provides 3D semantic segmentation annotations, including buildings, roads, cars, and trees, as well as 3D building-level instance annotations (see \Tref{tab:dataset}). 
We select four regions with a high density of building instances and various architecture styles, namely \emph{Yingrenshi}, \emph{Yuehai-Campus}, \emph{Longhua-1}, and \emph{Longhua-2}. 
We downsample images by four times for training and uniformly sample around ten images as the testing set for each scene.

\begin{table}[tb]\centering
    \caption{Statistics of the UrbanBIS \zyqm{dataset}~\cite{yang2023urbanbis}.}
    \label{tab:dataset}
    \resizebox{\linewidth}{!}{%
\begin{tabular}{cccccc}
\hline
Dataset & \zyqm{Covered} area & Number of images & Resolutions & Building instances \\
\hline
Yingrenshi & 440 $\times$ 220 $m^2$ & 854 & 3648 $\times$ 5472 & 41 \\
Yuehai-Campus & 900 $\times$ 280 $m^2$ & 955 & 3648 $\times$ 5472 & 30 \\
Longhua-1 & 550 $\times$ 530 $m^2$ & 999 & 5460 $\times$ 8192 & 26 \\
Longhua-2 & 550 $\times$ 300 $m^2$ & 677 & 5460 $\times$ 8192 & 38 \\
\hline
\end{tabular}%
}

    \vspace{0.1em} 
    \caption{Comparison with 2D segmentation methods.}
    \label{tab:2d_segmentation}
    \small
\resizebox{0.48\textwidth}{!}{
\begin{tabular}{l|ll|ll|ll|ll}  
\toprule
\multirow{2}{*}{Method} 
& \multicolumn{2}{c|}{Yingrenshi} 
& \multicolumn{2}{c|}{Yuehai-Campus} 
& \multicolumn{2}{c|}{Longhua-1} 
& \multicolumn{2}{c}{Longhua-2}
\\
& building  & road  
& building  & road  
& building  & road  
& building  & road 
\\
\midrule
UNetFormer~\cite{wang2022unetformer} 
& 76.0 & 17.5
& 68.7 & 24.5  
& 77.5 & 4.5 
& \textbf{78.4} & 10.8
\\
Mask2Former~\cite{cheng2022masked} 
& 84.8 & 49.7
& 70.9 & 44.7 
& 68.5 & 47.5  
& 70.0 & 42.5 
\\
Scale-adaptive fusion 
& \textbf{93.2} & \textbf{56.8}
& \textbf{90.7} & \textbf{52.0} 
& \textbf{77.9} & \textbf{48.7}
& 77.4 & \textbf{43.0}
\\
\bottomrule
\end{tabular}}
\end{table}

\begin{table*}[t] \centering
    \newcommand{\Frst}[1]{\textcolor{red}{\textbf{#1}}}
    \newcommand{\Scnd}[1]{\textcolor{blue}{\textbf{#1}}}
    \caption{Quanlitative comparison on the novel-view synthesis and the semantic segmentation on the UrbanBIS dataset \cite{yang2023urbanbis}. Mask2Former is a 2D segmentation method and cannot be evaluated for PSNR, and Semantic-NeRF (M2F) shares the same geometry with ours.}
    \label{tab:results_semantic}
    \small
\resizebox{\textwidth}{!}{
\begin{tabular}{l|llll|llll|llll|llll}  
\toprule
\multirow{2}{*}{Method} 
& \multicolumn{4}{c|}{Yingrenshi~\cite{yang2023urbanbis}}   
& \multicolumn{4}{c|}{Yuehai-Campus~\cite{yang2023urbanbis}} 
& \multicolumn{4}{c|}{Longhua-1~\cite{yang2023urbanbis}} 
& \multicolumn{4}{c}{Longhua-2~\cite{yang2023urbanbis}}
\\
& mIoU$\uparrow$ & Building$\uparrow$ & Car$\uparrow$ & PSNR$\uparrow$ 
& mIoU$\uparrow$ & Building$\uparrow$ & Car$\uparrow$ & PSNR$\uparrow$  
& mIoU$\uparrow$ & Building$\uparrow$ & Car$\uparrow$ & PSNR$\uparrow$ 
& mIoU$\uparrow$ & Building$\uparrow$ & Car$\uparrow$ & PSNR$\uparrow$
\\
\midrule
Mask2former~\cite{cheng2022masked} 
& 58.4 & 84.8 & 28.4 & --
& 57.9 & 70.9 & 42.3 & --
& 49.2 & 68.5 & 20.7 & -- 
& 46.2 & 70.0 & 23.5 & --
\\
Panoptic-Lift~\cite{siddiqui2023panoptic} 
& 32.9 & 83.0 & 1.4 & 21.2 
& 33.7 & 69.0 & 0.1 & 21.5
& 29.5 & 60.2 & 0.2 & 21.9
& 20.4 & 59.9 & 0.2 & 20.9
\\
Semantic-NeRF (M2F)~\cite{zhi2021place} 
& 67.6 & 92.9 & 39.5 & -- 
& 70.9 & 88.1 & 45.3 & --
& 61.1 & 86.4 & 34.9 & -- 
& 64.1 & 89.9 & 37.5 & --
\\
Ours
& \textbf{72.0} & \textbf{95.5} & \textbf{49.3} & \textbf{26.8}
& \textbf{74.9} & \textbf{94.1} & \textbf{46.2} & \textbf{29.8}
& \textbf{66.1} & \textbf{87.3} & \textbf{43.8} & \textbf{26.7} 
& \textbf{66.7} & \textbf{91.0} & \textbf{40.9} & \textbf{26.4}
\\
\bottomrule
\end{tabular}}

\end{table*}

\begin{figure*}[t] \centering
    \input{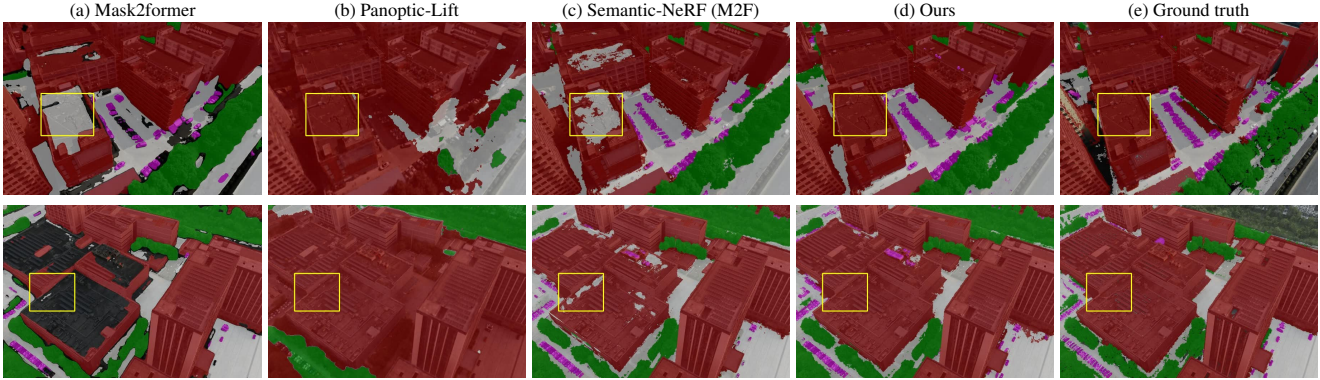}
    \caption{Qualitative comparison of semantic segmentation on \emph{Yingrenshi} and \emph{Longhua-2} from the UrbanBIS dataset (\emph{Building}: Red, \emph{Road}: White, \emph{Car}: Violet, \emph{Tree}: Green, unrecognized areas of Mask2Former: Black). Areas without masks in (e) have no GT annotation.}
    \label{fig:qualitative}
\end{figure*}

\paragraph{Evaluation Metrics}
We measure the quality of the novel-view synthesis and semantic segmentation in terms of PSNR and the mean intersection over union (mIoU), respectively. 
To evaluate the instance building segmentation, we use a scene-level Panoptic Quality (PQ$^\text{scene}$) metric~\cite{siddiqui2023panoptic}, which takes the consistency of the instance across different views into account. As we focus on the segmentation of building instances, we report the PQ$^\text{scene}$ of the building.

\subsection{Evaluation on Semantic Segmentation}%
\label{sub:Results I}
\paragraph{Choice of 2D Segmentation Method}
We discuss two types of 2D segmentation methods for providing the semantic labels for each aerial image, namely the UNetFormer~\cite{wang2022unetformer}, which is designed for the aerial images semantic segmentation, and Mask2Former~\cite{cheng2022masked}, which is a universal panoptic segmentation method.
 As shown in \Tref{tab:2d_segmentation}, UNetFormer does not generalize well on the UrbanBIS dataset, which fails to segment the \emph{Road}. 
Therefore, we take Mask2Former as the foundation to get the initial 2D semantic segmentation. 

\paragraph{2D Semantic Label Fusion} 
We first demonstrate the effectiveness of the proposed \semanticfusion by evaluating the accuracy of the 2D semantic labels. 
We can see from \Tref{tab:2d_segmentation} that the proposed \semanticfusionshort can significantly improve the accuracy, especially for the building \zyqm{category}.

\begin{figure*}[t] \centering
    \input{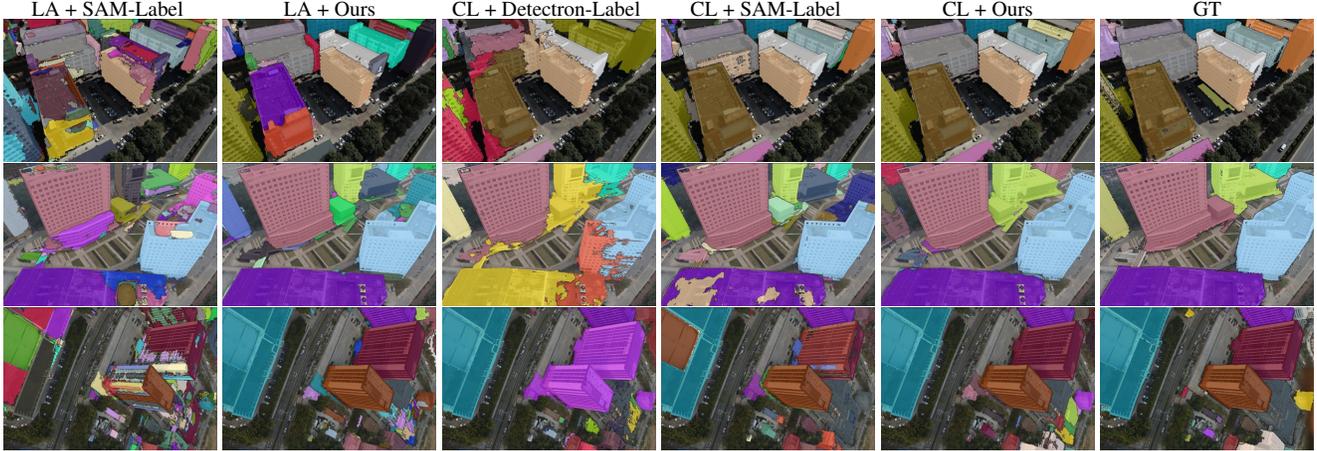}
    \caption{Qualitative comparison on the building instance segmentation. From top to bottom, we show the results of different approaches in three scenarios: \emph{Yingrenshi}, \emph{Yuehai-Campus}, and \emph{Longhua-2}. Different instances are represented in different colors.} 
    \label{fig:qualitative_instance}
\end{figure*}

\paragraph{3D Semantic Field}
To evaluate the performance of semantic segmentation in 3D lifting, we conducted a comparative analysis of our method against the official implementation of Panoptic-Lift~\cite{siddiqui2023panoptic}, and a modified Semantic-NeRF \cite{zhi2021place}. 
Panoptic-Lift employs the universal Mask2Former for predicting 2D semantic labels and lifts 2D labels to 3D for room-scale scene. 
However, Panoptic-Lift performs worse on the urban scene due to the worse geometry reconstruction as it employs the TensoRF~\cite{chen2022tensorf} as the backbone, which struggles to scale up to a high grid resolution.
For a fair comparison, we designed a variant of semantic-NeRF using the same geometry backbone as ours (\ie, high-resolution \hashgrid). This modified semantic-NeRF is trained with semantic labels from Mask2Former, while our method is trained with fused labels.

Quantitative results in Table~\ref{tab:results_semantic} demonstrate that our method outperforms the others in terms of mIoU through the use of \semanticfusionshort, highlighting its effectiveness. Moreover, \Fref{fig:qualitative} shows that our method solves the issue of misclassification resulting from the ambiguity between building roofs and road surfaces in aerial images, leading to better results.

\subsection{Evaluation on Instance Building Segmentation}%
\label{sub:Results II}
To lift 2D instance labels to 3D, we utilize two different optimization methods: the linear assignment from Panoptic-Lift~\cite{siddiqui2023panoptic} and the contrastive learning from Contrastive-Lift~\cite{bhalgat2023contrastive} which \zyqm{is} followed by HDBSCAN~\cite{mcinnes2017hdbscan} as post-processing cluster algorithm. Moreover, we did not make a comparison with the Panoptic-Lift official implementation because of its poor semantics results. 
To mitigate the impact of geometry reconstruction and semantic segmentation, we employ the same geometry and semantic results across experiments in this section.

\begin{table}[tb]\centering
    \caption{Quantitative comparison of instance segmentation in PQ$^\text{scene}$ of building category. For brevity, LA and CL denote the linear assignment and contrastive learning, respectively.}
    \label{tab:results_instance}
    \small
\resizebox{0.48\textwidth}{!}{
\begin{tabular}{l|cccc}  
\toprule
{Method}
& Yingrenshi
& Yuehai-Campus
& Longhua-1
& Longhua-2
\\
\midrule
LA + \detectronlabel~\cite{spacenet2023}
& 15.8
& {40.8}
& {38.2}
& 18.2
\\
LA + \SAMlabel~\cite{kirillov2023segany}
& 14.4
& 4.0
& 4.5
& 4.0
\\
LA + Ours
& {38.7}
& 26.0
& 36.3
& {19.0}
\\
\midrule
CL + \detectronlabel~\cite{kirillov2023segany}
& 26.6
& {30.6}
& 36.5
& 17.3
\\
CL + \SAMlabel~\cite{kirillov2023segany}
& 54.8   
& {18.8}
& 29.7
& 22.9
\\
CL + Ours
& {\textbf{64.1}}
& {\textbf{43.6}}
& {\textbf{45.8}}
&   \textbf{31.5}
\\
\bottomrule
\end{tabular}}
\end{table}

We establish two baselines, one trained with instance labels obtained from the Detectron~\cite{spacenet2023} and one with labels from SAM~\cite{kirillov2023segany}.
For brevity, we refer to these baselines as \detectronlabel and \SAMlabel. 
Our method builds upon \SAMlabel by incorporating the cross-view label grouping. 

\Tref{tab:results_instance} presents the PQ$^\text{scene}$ metric on four urban scenes.
Results on \emph{Yingrenshi} reveal that training with \detectronlabel struggles with dense building instances due to inaccurate instance segmentation. 
While the \SAMlabel model achieves reasonable results in \emph{Yingrenshi}, it struggles to handle large buildings with diverse shapes as shown in the other three \zyqm{scenes}.
It is because SAM tends to produce over-segmented labels for these buildings, leading to cluttered 3D segmentation, particularly trained with linear assignment.
With our proposed cross-view label grouping strategy, we significantly improve the performance compared to that trained with \SAMlabel in both linear assignment and contrastive learning. The best results are achieved by integrating the \instancefusionshort with the contrastive learning. \Fref{fig:qualitative_instance} illustrates the qualitative comparison, where our approach exhibits more accurate segmentation results, further affirming the effectiveness of our method.

\subsection{Ablation Analysis}%
To further verify the design of our method, we conduct ablation studies on the \emph{Yingrenshi} dataset.

\paragraph{Effect of Instance Label Grouping}
We evaluate the effectiveness of the components employed in the proposed instance segmentation method. As depicted in \Tref{tab:ablation_geo}, the utilization of geometry-guided instance filtering can improve instance segmentation in some extent, compared to the baseline trained with \SAMlabel. 
More importantly, utilizing the cross-view grouping strategy achieves significant improvement in instance segmentation, demonstrating the effectiveness of the \instancefusionshort strategy.

\begin{table}[tb]\centering
    \vspace{-1em}
    \caption{Effect of cross-view label grouping (PQ$^\text{scene}$).}
    \label{tab:ablation_geo}
    \small
\resizebox{0.48\textwidth}{!}{
\begin{tabular}{l|ccc}  
\toprule
{Method}
& Linear assignment
& Contrastive learning
\\
\midrule
Baseline (\SAMlabel) & 14.4   &54.8  
\\
Baseline + Filter    &19.8  & 55.9  
\\
Baseline + Filter + Cross-view  &\textbf{38.7} & \textbf{64.1}
\\
\bottomrule
\end{tabular}}
    \vspace{0.1em}
    \caption{Effect of geometry reconstruction quality.}
    \label{tab:depth_prior}
    \small
\resizebox{0.48\textwidth}{!}{
\begin{tabular}{l|cccccc}  
\toprule
{Method} 
&{PSNR} 
&{mIoU} 
&{Building} 
&{Road}
&{Car}
&{Tree}
\\
\midrule
Panoptic-Lift~\cite{siddiqui2023panoptic} 
&21.24  &32.9 	&83.0 	&32.6 	&1.4 	&14.5 
\\
Ours without depth-prior 
&25.01  &70.4 	&94.9 	&66.6 	&46.7 	&73.3  \\
Ours with depth-prior
&\textbf{26.79} &\textbf{72.0} 	&\textbf{95.5} 	&\textbf{68.9} 	&\textbf{49.3} 	&\textbf{74.5} 
\\
\bottomrule
\end{tabular}}
\end{table}

\begin{figure}[t] \begin{center}
\vspace{-1em}
    \includegraphics[width=0.48\textwidth]{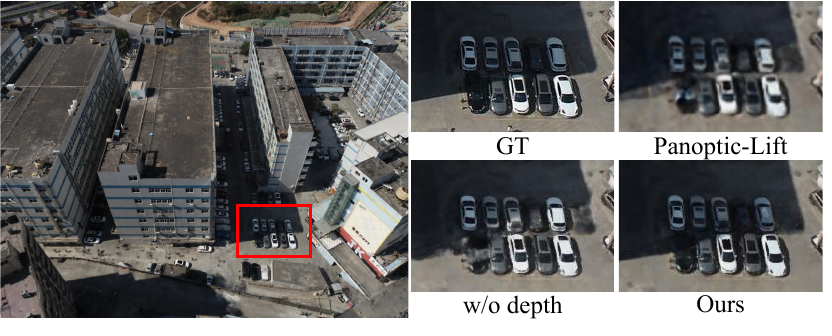}
    \caption{Visualization of novel-view synthesis.} 
    \label{fig:depth_prior}
    \vspace{-2.0em}
\end{center} \end{figure}

\paragraph{Effect of Geometry Reconstruction}
To investigate the effect of the geometry reconstruction, we compare the novel-view synthesis and semantic segmentation results in \Tref{tab:depth_prior}. We can see from the table that Panoptic-Lift suffers from low-quality reconstruction, resulting in poor segmentation of small objects (\eg car and tree categories). By incorporating the depth-prior from multi-view stereo, the rendering quality and segmentation quality can be effectively improved. \Fref{fig:depth_prior} shows an example of the novel-view synthesis.

\section{Conclusion}
\label{sec:Conclusion}
In this paper, we have introduced a neural radiance field method for urban-scale semantic segmentation and building-level instance segmentation from aerial images. Our method lifts noisy 2D labels, predicted by off-the-shelf methods, to 3D without manual annotations. 
We proposed a \semanticfusion strategy that significantly improves the segmentation results across objects of varying sizes. 
To achieve multi-view consistent instance supervision for building instance segmentation, we introduced a \instancefusion strategy based on the 3D scene representation. 
In addition, we enhanced the reconstructed geometry by incorporating the depth prior from multi-view stereo, leading to more accurate segmentation results. 
Experiments on multiple real-world \zyqm{scenes} demonstrate the effectiveness of our method.

\paragraph{Future Work} 
Currently, our method focuses on close-vocabulary scene understanding. Recent methods have shown promising results in open-vocabulary understanding by distilling CLIP features into NeRF \cite{kerr2023lerf,kobayashi2022decomposing}. 
Nonetheless, feature conflicts caused by varying object sizes and multi-view inconsistency can impair the distillation. In the future, 
we aim to apply the proposed method to enhance the feature distillation process.

\paragraph{Acknowledgement.} 
\zyqm{This work was supported in part by NSFC with Grant No.~62293482, the Basic Research Project No.~HZQB-KCZYZ-2021067 of Hetao Shenzhen-HK S\&T Cooperation Zone. It was also partially supported by NSFC with Grant No.~62202409, Shenzhen Science and Technology Program with Grant No.~RCBS20221008093241052, the National Key R\&D Program of China with grant No.2018YFB1800800, by Shenzhen Outstanding Talents Training Fund 202002, by Guangdong Research Projects No.2017ZT07X152 and No.2019CX01X104, and by the Guangdong Provincial Key Laboratory of Future Networks of Intelligence (Grant No.2022B1212010001).
}

{\small
\bibliographystyle{ieee_fullname}
\bibliography{ref_largescale}
}

\clearpage

\setcounter{figure}{0}
\setcounter{table}{0}
\begin{appendices}
\startcontents[supple]
{
    \hypersetup{linkcolor=black}
    \printcontents[supple]{}{1}{}
}
\renewcommand{\thesection}{\Alph{section}}%
\renewcommand\thetable{\Roman{table}}
\renewcommand\thefigure{\Roman{figure}}

\section{More Details for the Method}

\paragraph{Implementation Details} We implemented our method with PyTorch~\cite{paszke2017pytorch} and used the Adam optimizer~\cite{kingma2014adam} with a learning rate of 0.001 for the {\hashgrid} and 0.01 for the semantic and instance MLPs. 
\zyqm{We combine the tri-plane features~\cite{zhang2023efficient} and a cuboid hash-grid proposed by StreetSurf~\cite{guo2023streetsurf}, as a backbone for geometry reconstruction.}
The {\hashgrid} was trained with a hash level of $L=16$, the highest resolution of $R=8192$, and a hash table size of $T=2^{22}$. The architectures of the semantic and instance networks are identical, each consists of a 5-layer MLP with 128 channels. \zyqm{Moreover, for a scene represented by a volume of [0, 1], we raise the altitude of all cameras by displacing each camera in the opposite direction of the camera's focal point with an offset of 0.3, during the \semanticfusion. The geometry-guided instance filtering threshold is empirically set to 10 meters (in physical space) for all testing scenes.}

\subsection{\InstanceFusion}
\Fref{fig:supp_grouping} illustrates cases of \InstanceFusion on the \textit{Longhua-1} and \textit{Yingrenshi} datasets. 
For example, in the SAM instance label, mask blocks A, B, and C belong to the same building but are segmented as three distinct instances, introducing ambiguity in the supervision label during training. This can result in two pixels from the same building being incorrectly labeled as different instances. 
In contrast, with our \instancefusion, separated blocks of the same instance are merged into the same group (e.g., group1: \{A, B, C\}; group2: \{D, E\}), guiding the training of the instance field more effectively.
Specifically, during training, we randomly select a single instance from each group (e.g., blocks A and D in the SAM label) to reduce the occurrence of conflicts. 
The pseudo-code for the proposed \instancefusion is shown in Algorithm \ref{alg:instancefusion}.

\begin{figure}[t] 
    \begin{center}
    \includegraphics[width=0.4\textwidth]{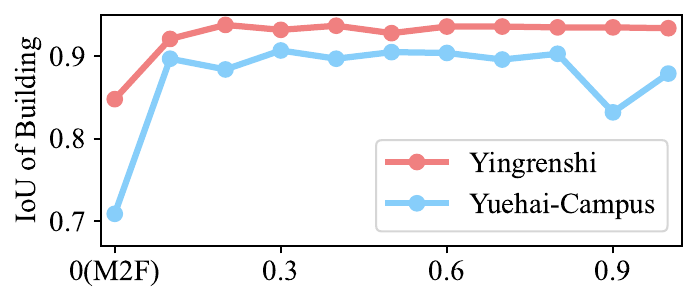}
    \captionof{figure}{\zyqm{Effect of altitude offset in semantic fusion strategy.}}
    \label{fig:altitude}
    \end{center} 
    \begin{center}
    \includegraphics[width=0.4\textwidth]{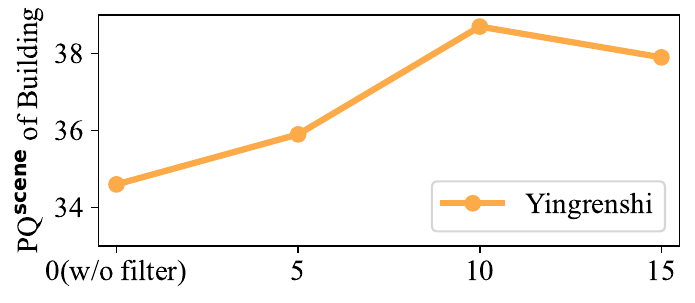}
    \captionof{figure}{\zyqm{Effect of geometry-guided instance filtering threshold.}}
    \label{fig:geo}
    \end{center} 
 \end{figure}

\subsection{Mask2Former Semantic Label Mapping}%
\label{par:Mask2Former label mapping}
Following Panoptic-Lift~\cite{siddiqui2023panoptic}, we employ the universal 2D segmentation method, Mask2Former~\cite{cheng2022masked}, to obtain semantic labels and utilize the implementation\protect\footnote{An implementation of Mask2former with test-time augmentation: \url{https://github.com/nihalsid/Mask2Former}.} with test-time augmentation.
The original Mask2former provides pre-trained models on various datasets, including COCO~\cite{lin2014microsoft}, Cityscapes~\cite{cordts2016cityscapes}, ADE20K~\cite{zhou2017scene}, \etal. We observed that the model trained on the ADE20K dataset (\text{swin\_large\_IN21k} model) demonstrates robust performance for semantic segmentation of aerial images. For training, we map the ADE20K classes (150 classes in total) into four categories: Building, road, car, and tree. Additionally, we marked the category from the 150 classes that may not appear in aerial images as \textit{Cluster} (\eg, indoor objects), mitigating interference from inaccurate segmentation results of Mask2former.

\zyqm{Moreover, during the processing of the \semanticfusion, the images with the original size are cropped into four parts to obtain segmentation results, as feeding the entire image may lead to out-of-memory errors. Then, the car and tree segmentation results from the downscaled images of Mask2Former are substituted.}

\begin{figure*}[t] \centering
    \input{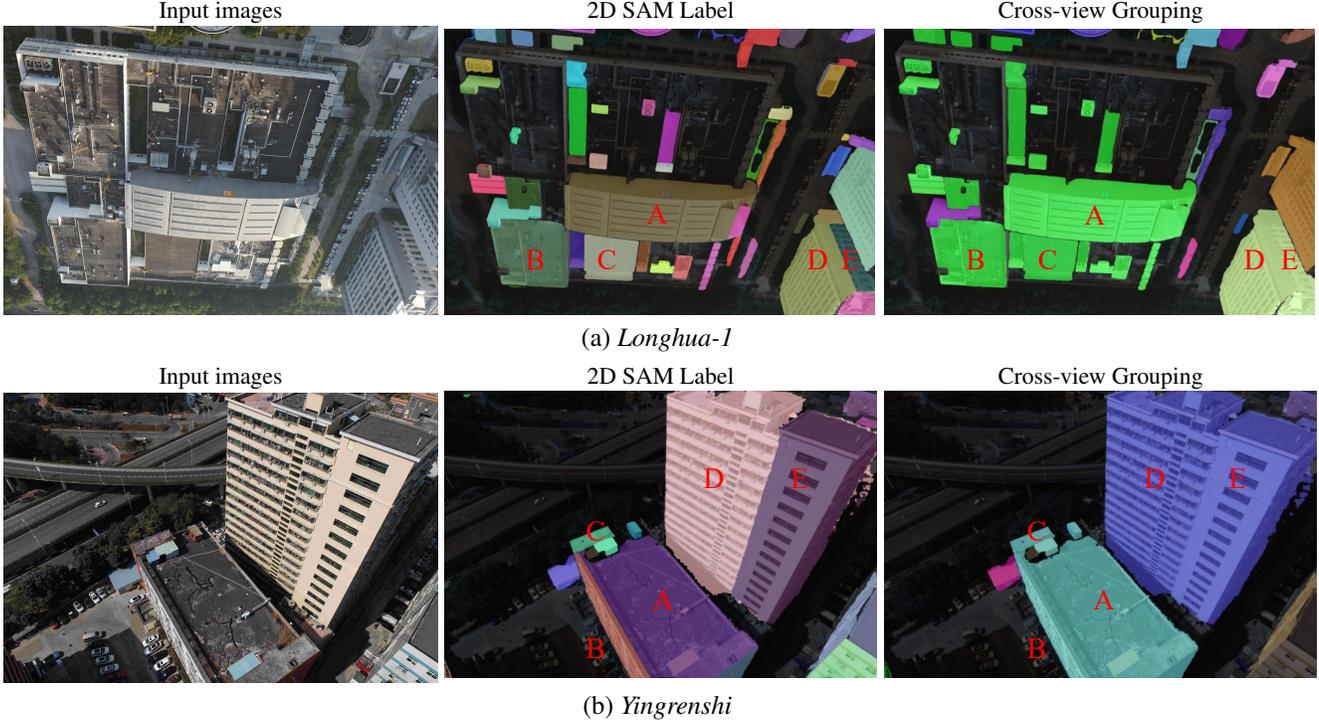}
    \caption{Illustration of \InstanceFusion. Different colors represent different instances. SAM~\cite{kirillov2023segany} produces over-segmented masks and an instance might be segmented into different blocks (\eg, A, B, and C belong to the same building but are incorrectly divided into different instances). Our \instancefusion alleviates this problem, reducing the conflict of 2D instance supervision during training.} 
    \label{fig:supp_grouping}
\end{figure*}

\begin{algorithm*}
\caption{Pseudo code for the \instancefusion strategy.}
\label{alg:instancefusion}
    \setlength{\algorithmicindent}{2em}
\begin{algorithmic}[1] %
    \REQUIRE $N$ images with SAM masks $\msam_i$ for each $i$-th view
    \ENSURE Cross-view guidance map $U_i$ for each view

    \FOR{$i = 1$ to $N$}
        \STATE Project SAM masks from all other views onto the $i$-th view: $\{\msam_{j\to i} | j = 1, \dots, N, j \neq i\}$
        \FOR{each instance mask $\msam_i^k$ in the $i$-th view}
            \FOR{each instance mask $\msam_{j\to i}^l$ in projected mask $\msam_{j\to i}$}
                \IF{$\frac{|\msam_i^k \cap \msam_{j\to i}^l|}{\min(|\msam_i^k|, |\msam_{j\to i}^l|)} > \tau$}
                    \STATE expanded mask: $\msam_{i\cup j}^k$.append($\msam_{j\to i}^l$)
                \ENDIF
            \ENDFOR
            \STATE Combine all $\msam_{i\cup j}^k$ to form cross-view mask $U_i^k$: \hbox{$U_i^k = \bigcup_{j \neq i} \msam_{i\cup j}^k$}
            
        \ENDFOR
        \STATE Organize cross-view masks in ascending order based on area
        \STATE Sequentially layer cross-view masks onto map $H \times W$ to form $U_i$
    \ENDFOR
\end{algorithmic}

\end{algorithm*}

\subsection{Effect of Hyper-parameter}

\paragraph{Setting of far view in semantic fusion.} 
For a scene represented by a volume of [0, 1], we raise the altitude of all cameras by displacing each camera in the opposite direction of the camera's focal point with an offset of 0.3. 
\Fref{fig:altitude} shows that the scale-adaptive semantic fusion consistently enhances the results of Mask2Former and remains effective across various offset values.
This strategy is inspired by the observation that large object recognition benefits from a larger receptive field, leading to more reliable segmentation.

\paragraph{Geometry-guided instance filtering.}
The geometry-guided instance filtering threshold is empirically set to 10 meters (in physical space) for all testing scenes. 
\Fref{fig:geo} shows that applying the filtering can improve results \zyqm{against w/o filter}, and the filtering works effectively in the range of [5, 15] meters.

\section{Details for the Instance Field Optimization}
Our building instance segmentation method is built upon the 2D image segmentation method. In selecting the base model for 2D image segmentation, we considered SAM~\cite{kirillov2023segany} and Detectron~\cite{wu2019detectron2}. During our testing, we found that Detectron did not perform well on aerial images for building segmentation. Consequently, we experimented with Detectron2-SpaceNet~\cite{spacenet2023}, which is fine-tuned on the SpaceNet dataset~\cite{van2018spacenet} and based on the Mask-RCNN model from Detectron~\cite{wu2019detectron2}. While its segmentation performance showed improvement, it did not generalize well to diverse urban scenes (refer to Figure 3 of the main paper). Therefore, we decided to build our model upon the SAM model.

As stated in the paper, we utilize two methods to achieve the 3D building instance segmentation: \textit{linear assignment} from Panoptic-Lift~\cite{siddiqui2023panoptic} and \textit{contrastive learning} from Contrastive-Lift~\cite{bhalgat2023contrastive}.

\paragraph{Linear Assignment\protect\footnote{We utilize the official implementation from Panoptic-Lift: \url{https://github.com/nihalsid/panoptic-lifting}.}}%
2D machine-generated instance labels are noisy and view-inconsistent, for which Panoptic-Lift~\cite{siddiqui2023panoptic} proposes to map them into 3D surrogate identifiers, and finds out the most compatible injective mapping by solving a linear assignment problem. 
Let $U\left(\boldsymbol{r}\right)$ denotes the instance segment label of the pixel casting ray 
$\boldsymbol{r}$, $\boldsymbol{R}_k$ the subset of rays in $\boldsymbol{R}$ that belong to 2D instance $k\in K_\mathbf{I}$, and $K_R\subseteq K_\mathbf{I}$ the subset of 2D instances that are represented in the batch of rays $\boldsymbol{R}$, the optimal injective mapping is then given by:
\begin{equation}
    \Pi_{\boldsymbol{R}}^{\star}=\underset{\Pi_{\boldsymbol{R}}}{\operatorname{argmax}} \sum_{k \in K_{R}} \sum_{\boldsymbol{r} \in \boldsymbol{R}_k} \frac{\tilde{S}(\boldsymbol{r})_{\left(\Pi_\mathbf{I}(U(\boldsymbol{r}))\right)}}{\left|\boldsymbol{R}_k\right|}  
\end{equation}
where $\tilde{S}(\boldsymbol{r})_{\left(\Pi_\mathbf{I}(U(\boldsymbol{r}))\right)}$ denotes the $\Pi_\mathbf{I}(U(\boldsymbol{r}))$-th element of the instance label vector $\tilde{S}(\boldsymbol{r})$. 

Thus the instance loss can be formulated as follows:
\begin{equation}
    \mathcal{L}_{\text {instance}}=-\frac{1}{|\boldsymbol{R}|} \sum_{\boldsymbol{r} \in \boldsymbol{R}} w_r \log \tilde{S}(\boldsymbol{r})_{\left(\Pi_{\boldsymbol{R}}^{\star}\left(U(\boldsymbol{r})\right)\right)}    
\end{equation}
where $w_r$ is the prediction confidence.\\

\paragraph{Contrastive Learning\protect\footnote{We utilize the official implementation from Contrastive-Lift: \url{https://github.com/yashbhalgat/Contrastive-Lift}.}}%

Instead of aligning labels extracted from multiple views, Contrastive-Lift~\cite{bhalgat2023contrastive} directly learns embeddings from the noisy 2D machine-generated labels via optimizing a contrastive loss and acquires the instance segments by simply clustering the embeddings. The instance loss can be formulated as follows:
\begin{equation}
    \mathcal{L}_{\text {instance}}=\mathcal{L}_{\text {sf}}+\mathcal{L}_{\text {conc}}
\end{equation}
where the first item $\mathcal{L}_{\text {sf}}$ is the contrastive loss using a slow-fast learning scheme, and the second item $\mathcal{L}_{\text {conc}}$ is the concentration loss used to encourage the embeddings to form concentrated clusters for each object.

Specifically, given the two non-overlapping subsets $\boldsymbol{R}_1$and $\boldsymbol{R}_2$ partitioned from rays in $\boldsymbol{R}$, the contrastive loss function is:

\begin{equation}
\begin{aligned}
    & \mathcal{L}_{\text {sf}} = -\frac{1}{\left|\boldsymbol{R}_1\right|} \cdot \\
    & \sum_{\boldsymbol{r} \in \boldsymbol{R}_1} \log \frac{\sum_{\boldsymbol{r}^{\prime} \in \boldsymbol{R}_2} \mathbf{1}_{U(\boldsymbol{r})=U\left(\boldsymbol{r}^{\prime}\right)} \exp \left(\operatorname{sim}\left(\tilde{S}(\boldsymbol{r}), S({\boldsymbol{r^{\prime}}}) ; \gamma\right)\right)}{\sum_{\boldsymbol{r}^{\prime} \in \boldsymbol{R}_2} \exp \left(\operatorname{sim}\left(\tilde{S}(\boldsymbol{r}), S({\boldsymbol{r^{\prime}}}) ; \gamma\right)\right)}
\end{aligned}
\end{equation}
where $\mathbf{1}$ is the indication function, $\operatorname{sim}\left(x, x^{\prime} ; \gamma\right)=\exp \left(-\gamma\left\|x-x^{\prime}\right\|^2\right)$is used to compute the similarity between embeddings in Euclidean space,  and $S({\boldsymbol{r^{\prime}}})$ is the instance label inferred by the slowly-updated embedding field~\cite{bhalgat2023contrastive}. 

And the concentration loss function is:
\begin{equation}
    \mathcal{L}_{\text {conc}}=\frac{1}{\left|\boldsymbol{R}_1\right|} \sum_{\boldsymbol{r} \in \boldsymbol{R}_1}\left\|\tilde{S}(\boldsymbol{r})-\frac{\sum_{\boldsymbol{r}^{\prime} \in \boldsymbol{R}_2} \mathbf{1}_{U(\boldsymbol{r})=U\left(\boldsymbol{r}^{\prime}\right)} S({\boldsymbol{r^{\prime}}})}{\sum_{\boldsymbol{r}^{\prime} \in \boldsymbol{R}_2} \mathbf{1}_{U(\boldsymbol{r})=U\left(\boldsymbol{r}^{\prime}\right)}}\right\|^2
\end{equation}

For clustering, we use HDBSCAN~\cite{mcinnes2017hdbscan} clustering in experiments following Contrastive-Lift~\cite{bhalgat2023contrastive}. We sample all rays in the testing images and then randomly select 200000 pixels of building for clustering, with the minimum cluster size set to 1000.

\begin{figure*}[t] \centering
    \includegraphics[width=1\textwidth]{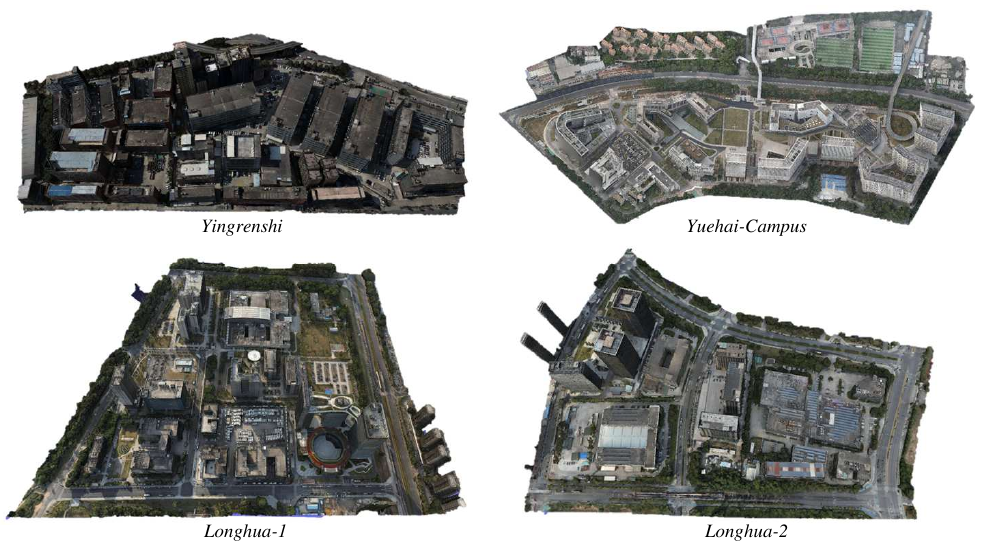}
    \caption{Bird's eye view of the \emph{Yingrenshi}, \emph{Yuehai-Campus}, \emph{Longhua-1} and \emph{Longhua-2} areas in UrbanBIS dataset~\cite{UrbanScene3D}.} 
    \label{fig:datasets bev}
\end{figure*} 

\section{More Details for the Dataset}

\subsection{Dataset Selection}
We evaluate our method on UrbanBIS dataset~\cite{yang2023urbanbis}, which provides 3D building-level instance annotations and 3D semantic segmentation annotations of six categories, including buildings, roads, cars, and trees. We select four regions with a high density of building instances and various architecture styles, namely \textit{Yingrenshi, Yuehai-Campus, Longhua-1, and Longhua-2}. \Fref{fig:datasets bev} shows the bird's eye view of the four mentioned areas, which are covered by a diverse range of architectural instances.

\subsection{Ground-truth Label for Evaluation}
2D ground-truth label for each view is acquired by projecting the 3D point cloud annotations.
Specifically, UrbanBIS dataset~\cite{yang2023urbanbis} provides 3D point cloud annotations for semantic and instance segmentation, along with 2D aerial images. However, individual 2D annotations for semantic and instance segmentation are not provided for each image, and camera poses for projections onto 2D images are also not available. Moreover, the point cloud given by UrbanBIS is sparse, resulting in unsatisfactory projections on 2D images.

To address these limitations, we reconstruct a dense point cloud and corresponding camera poses from the 2D aerial images. Subsequently, we register the reconstructed point cloud with the annotated UrbanBIS point cloud using CloudCompare and annotate our reconstructed points by employing KD-tree~\cite{bentley1975multidimensional} algorithm to find out the labels of the nearest annotated points in the UrbanBIS point cloud from ours. This process allows us to obtain annotations for the dense point cloud with known camera poses, which are then projected onto 2D images, yielding 2D image annotations.

It is important to note that we made modifications to the original annotations for two reasons. Firstly, the ground-truth annotations are not sufficiently accurate, mainly regarding missing annotations of cars. Secondly, for the Yuehai-Campus area, UrbanBIS has not provided corresponding 2D aerial images so far. We then utilized images from the UrbanScene dataset~\cite{UrbanScene3D}, which covers the Yuehai-Campus region but with a significant time gap compared to the UrbanBIS dataset~\cite{yang2023urbanbis}. 
Consequently, there are substantial discrepancies in the distribution of cars and trees between the UrbanBIS point cloud and our point cloud reconstructed from UrbanScene dataset~\cite{UrbanScene3D}. 
As modifying annotations on the point cloud would be time-consuming, we use the labeling tool ISAT~\cite{ISAT} to fix the 2D testing image annotations.

 \begin{figure}[t]

    \centering
    \makebox[0.15\textwidth]{\scriptsize Semantic-NeRF(M2F)}
    \makebox[0.15\textwidth]{\scriptsize Ours}
    \makebox[0.15\textwidth]{\scriptsize GT}
    \\
    \includegraphics[width=0.15\textwidth]{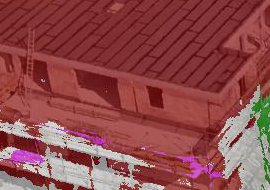}
    \includegraphics[width=0.15\textwidth]{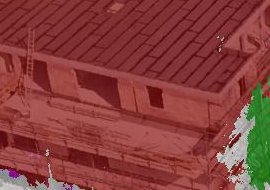}
    \includegraphics[width=0.15\textwidth]{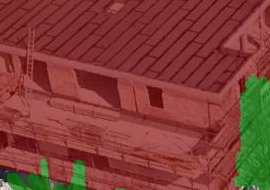}
    \captionof{figure}{Visual results on UAVid dataset.}
    \label{fig:uavid}
    
    \centering
    \captionof{table}{Comparison on UAVid dataset.}
    \label{tab:uavid}
    \small
\resizebox{0.48\textwidth}{!}{
\begin{tabular}{l|cc|cc}  
\toprule
\multirow{2}{*}{Method} 
& \multicolumn{2}{c|}{Sequence \#14} 
& \multicolumn{2}{c}{Sequence \#31} 
\\
& mIoU & Building 
& mIoU & Building
\\
\midrule
Mask2former 
& 64.9 & 74.9 & 57.8 & 73.3 
\\
Semantic-NeRF (M2F)
& 69.7 & 91.5 & 58.0 & 87.5 
\\
Ours
& \textbf{74.1} & \textbf{92.8} & \textbf{61.9} & \textbf{88.8} 
\\
\bottomrule
\end{tabular}}
    
\end{figure}

\section{More Results}
\subsection{Semantic Segmentation on UAVid Dataset}
\zyqm{To further evaluate the effectiveness of our method, we conduct experiments on the UAVid dataset, which contains 2D semantic labels for sparse frames.}
We conducted additional evaluations of semantic segmentation by choosing two video sequences for which the camera trajectory has a wide coverage area and can be reconstructed using COLMAP. 
The results presented in \Tref{tab:uavid} and \Fref{fig:uavid} highlight that our method outperforms the baseline methods, demonstrating the effectiveness of our semantic fusion strategy. %

\subsection{\zyqm{Comparison with Point-based Method}}
As there are no existing NeRF methods for aerial understanding, we adapt Semantic-NeRF to have the same backbone as ours to have a fair comparison.
To further validate the effectiveness of our method, we compare it with the SOTA point-based method~[31] on point cloud segmentation (during inference, the input is the GT point cloud). 
The model is trained on the SensatUrban dataset~\cite{hu2022sensaturban}.
For our method, we query the 3D point coordinates in the semantic field to obtain its predicted category.  
\Tref{tab:sota} shows that our method achieves more accurate results.

\begin{table}[t]\centering
    \resizebox{0.48\textwidth}{!}{
    \begin{tabular}{l|cccc}  
    \toprule
    {Method}
    & Yingrenshi & Yuehai-Campus & Longhua-1 & Longhua-2  \\
    \midrule
    RandLA-Net~\cite{hu2020randla}
    & 42.7 & 39.4 & 50.2 & 54.7    \\
    Ours
    & \textbf{62.9} & \textbf{55.7} & \textbf{66.5} & \textbf{66.0}     \\
    \bottomrule
    \end{tabular}}
    \captionof{table}{\zyqm{Comparison with the point-based method. The reported values are 3D mIoU.}}
    \label{tab:sota}
    \vspace{-1.5em}
\end{table}

\begin{figure*}[t] \centering
    \input{figs/qualitative_supp}
    \caption{Qualitative comparison of semantic segmentation on \emph{Yuehai-Campus} and \emph{Longhua-1} from UrbanBIS dataset (\emph{Building}: Red, \emph{Road}: White, \emph{Car}: Violet, \emph{Tree}: Green, unrecognized areas of Mask2former: Black). The areas without masks have no GT annotation in (e). Moreover, we present the novel-view synthesis results of our method.}
    
    \label{fig:qualitative_supp}
\end{figure*}

\subsection{More \zyqm{Visualization} Results}
\Fref{fig:qualitative_supp} shows more qualitative semantic segmentation results on the UrbanBIS dataset.

\section{Limitation}
Our method relies on a pre-trained 2D segmentation model and the SAM model to generate 2D labels. The failure of 2D methods will affect the final results. \zyqm{Moreover, our method needs a per-scene optimization for scene parsing.}

\stopcontents[supple]

\end{appendices}

\end{document}